\documentclass[accepted]{uai2024} 
                        
\usepackage[american]{babel}

\usepackage{natbib} 
\usepackage{mathtools} 
\usepackage{booktabs} 

\usepackage{tabularx}
\usepackage{csvsimple}
\usepackage{physics}
\usepackage{amsmath,amssymb,amsfonts}
\usepackage{algorithmic}
\usepackage{algorithm2e}
\usepackage{graphicx}
\graphicspath{ {Graphs/}{Table/}{New Graphs/}{Table_Graphs/} }
\usepackage{textcomp}
\usepackage{resizegather}
\usepackage{subfigure}
\usepackage{multicol}
\usepackage{multirow}
\usepackage{xcolor}
\usepackage{comment}
\usepackage{amssymb}
\usepackage{tikz}

\title{Towards Modeling Data Quality and Machine Learning Model Performance}

\author[1]{Usman Anjum}
\author[1]{Chris Trentman}
\author[1]{Elrod Caden}
\author[1]{Justin Zhan}
\affil[1]{%
    Computer Science Dept.\\
    University of Cincinnati\\
    Cincinnati, Ohio, USA
}
  
\begin{document}
\maketitle

\begin{abstract}
Understanding the effect of uncertainty and noise in data on machine learning models (MLM) is crucial in developing trust and measuring performance. In this paper, a new model is proposed to quantify uncertainties and noise in data on MLMs. Using the concept of signal-to-noise ratio (SNR), a new metric called deterministic-non-deterministic ratio (DDR) is proposed to formulate performance of a model. Using synthetic data in experiments, we show how accuracy can change with DDR and how we can use DDR-accuracy curves to determine performance of a model.
\end{abstract}

\section{Introduction}

Machine learning-based models and algorithms are being used widely for many different tasks. They are increasingly being used in sensitive domains like healthcare, finance \& banking, security, etc. to make critical decisions. In addition, most of the machine learning algorithms are "black-box" models which means the information of the internal workings is not known. As a result, we may not completely trust a machine-learning algorithm and its results \cite{kaplan2023trust} and it becomes necessary to understand the strength and limitations of a model. 

Biases, strengths, and limitations of machine learning algorithms can also arise from the quality of the data and inductive bias of the algorithm itself. In the machine learning domain, biases have also been referred to as "fairness". Fairness is the \textit{absence of any prejudice or favoritism towards an individual or group based on their inherent or acquired characteristics} \cite{mehrabi2021survey}. An unfair algorithm and low-quality data can result in skewed results that can lead to incorrect conclusions \cite{pagano2023bias}.  

A major reason for the data quality is the uncertainty and non-deterministic nature of data which is caused by the data source, and during data entry or acquisition. The non-deterministic nature of the data is usually considered as noise, but it may not necessarily be noise and there may be inherent properties of the data that are necessary for the analysis. For example, consider weather data that has some uncertainty due to patterns. Non-deterministic data would cause models to have difficulty in understanding and interpreting such data correctly. Most of the current research has focused on how to deal with the problem of non-deterministic nature of data sets. For example, filtering out noise \cite{quinlan1983learning, quinlan1986induction} or enhance the quality by pre-processing by 'hand' or automatically \cite{zhu2004class}. There are numerous pre-processing methods to deal with the noise and non-deterministic nature of data, such as eliminating the noisy instances completely and interpolation to correct the non-deterministic data to match the deterministic data points. However, the pre-processing techniques are dependant on the task being performed. Hence, the non-deterministic nature of data can adversely affect the results of any data analysis and reduce accuracy of the model.

Understanding the effect of noisy and nondeterministic data in the model can help understand the performance and limitations of a model under uncertain and noisy conditions. Uncertainty in data is an important aspect to consider when measuring model performance. Uncertainty can arise due to imperfect or unknown information \cite{hariri2019uncertainty}. Hence, if we can quantify and measure the uncertainty and noise, we could in fact get a true measure of the performance of a model. 

Current machine learning algorithms only focus on measuring the accuracy of the model, as to how well the predicted data match the actual data \cite{flach2019performance}. However, this information may not be enough, especially when we want to measure the performance of a model \cite{brown2021principles}. In fact a research conducted by \cite{wu2021current} explored the quality of data in anomaly detection and showed that the benchmark dataset used for anomaly detection had very little uncertainty in the data and hence, even the most simple of algorithms performed very well. The results from the paper showed that data quality also needs to be considered when measuring the performance of a model.

In this paper, we propose a framework that looks at quantifying data quality by as a function of uncertainty in the data set. In particular, we look at how uncertainty and non-deterministic data can effects accuracy in different model for different tasks and find the correlation between non-deterministic component in a data and the accuracy of the model. We focus on regression and classification tasks as non-determinism effects these tasks in different ways.

We begin by defining any data consisting of deterministic and non-deterministic parts. The non-deterministic part of a data can be noise or be present due to stochastic nature of data. The deterministic part of the data is defined by a definitive function. Consequently, we propose a new metric called the Deterministic-Non-Deterministic Ratio (DDR). DDR is basically the ratio of ``magnitude" of deterministic and non-deterministic portions of a data. The higher the DDR, the data have a higher deterministic part and consequently a model will be more accurate and able to make predictions. The lower the DDR, the data have a higher non-deterministic part and the model will make less accurate predictions.

We prove this hypothesis by running experiments on different models and tasks using synthetic data. We look at the task of regression and classification. Through these experiments we draw DDR-accuracy plots. Using the DDR-accuracy plot, we can determine the performance of a model.

The research questions that we aim to answer are:

\begin{itemize}
\item 
How does the score of a metric appropriate for a given type of machine learning model and calculated from the target values of a data set appropriate for that type of model and the predictions of the model based on the matrix / vector of feature values of that data set depend on the feature (vector) estimated from the matrix / vector of feature values of that data set?  
\item 
How does the performance of a machine learning model type in a data set appropriate for that type of model depend on both the feature (vector) of the matrix / vector of feature values of that data set and the value of a metric appropriate for that model type and calculated from the target values of that data set and the predictions of the model estimated from the matrix / vector of feature values of that data set? 
\end{itemize}

In summary, the main contributions of our paper are as follows: (1) we propose a new metric that can be used to quantify quality of data; (2) We formulate a model that explores the uncertainty and the determinism in data and its effect on accuracy; and (3) We use this model to measure a model's performance and conduct different experiments on a variety of tasks and show how our model can be used to show the applicability of our model.

\section{Related Works} \label{rel}

The need to develop a comprehensive technique to measure the performance of a model has been mentioned widely in previous literature \cite{brown2021principles}. Currently, the modern way to measure performance is to measure the difference between predicted and actual values. But these methods of measuring performance do not include data quality and the uncertainty.

One way of measuring the performance is to use the concept of trust. There are few ways that trust in AI can be quantified. 
Previous research has understood the importance of noise and its effect on the accuracy of a model. For example, the National Institute of Science and Technology (NIST) has modeled trust in the system in terms of two important parts. User Trust Potential, which is a function of the user, and Perceived System Trustworthiness, which is a function of the user, the system, and context \cite{stanton2021trust}. Consequently, the main attributes that define AI system trustworthiness are: Accuracy, Reliability, Resiliency, Objectivity, Security, Explainability, Safety, Accountability, and Privacy. Trust was measured as the user's perception of each attribute.

An example metric for measuring trust in a machine learning method was proposed by Wong et al. \cite{wong2020much}. The measure of trustworthiness of a particular deep neural network is according to how it operates under correct and incorrect result situations; trust density, a description of the distribution of general trust of a deep neural network for a particular answer situation; trust spectrum, a model of general trust in accordance with the variety of possible answer situations across both correctly and incorrectly answered questions; and NetTrustScore, a scalar metric summing up the general trustworthiness of a deep neural network according to the trust spectrum.

Another aspect of trust is in relation to how the model interacts with different qualities of data. The quality of data is related to the uncertainty in the data, mainly arising during data collection  \cite{hariri2019uncertainty}. This quality of the data is based on some features, such as noise, fairness, or bias, found in the data. A model that cannot handle low-quality data would be considered low in performance and less trustworthy than other models. Mehrabi et al. looked at the quality of data in terms of fairness and produced a taxonomy of different fairness definitions that other researchers made to avoid bias and compiled the results to show the unfairness present in current AI systems and how researchers address these issues \cite{mehrabi2021survey}. 

Quality in data was also studied by the papers in \cite{anjum2022localization} and \cite{anjum2023tbam}. According to the authors the data source determines the quality of the data and they define data quality as the reliability of the source and the amount of control over the data slate of the source.

Most current learning methods evaluate performance in terms of size of a randomly picked dataset on which they work, a metric termed sample complexity, that emphasizes quantity over quality. However, by the No-Free-Lunch theorem, quantity sometimes does not correlate with quality \cite{shalev2014understanding}. A second metric for the value of a data is its margin. The margin is defined as the smallest Euclidean distance between the convex hulls of the positive and negative elements of the data set \cite{shalev2014understanding}.  

Raviv et al. define the quality of the data set as the expected disagreement between a pair of random hypotheses from the set of hypotheses agreeing on the data set, a metric they term the expected diameter \cite{raviv2020value}. They showed that two datasets with equal margins and drastically different sets of consistent hypotheses can give identical results. Tosh et al. also used diameter to define the value of a dataset. In their case the diameter of a set of hypotheses was the maximum distance over all pairs of hypotheses in the set of hypotheses. The hypothesis distance was induced by the distribution over the data space between two hypotheses and is the probability under the distribution over the data space of the outputs of the two hypotheses being equal \cite{tosh2017diameter}.

Vapnik et al. propose a machine learning paradigm that reframes a problem from machine learning as a problem of estimating the conditional probability function as opposed to the problem of searching for the function that minimizes a given loss functional. This paradigm accounts for the relationship among elements of the data set and is therefore associated with data quality \cite{vapnik2019rethinking}.

We also would like to mention the work in \cite{Jiang2018trust} in which the authors have proposed a trust score for classifiers. They attempted to provide a more accurate trust metric than the basic confidence metrics typically offered by models that is calculated with some independence from the original model's confidence scoring. They look at the trust score of certain models and then observe whether they got correct results for a high trust score model or incorrect results for a low trust score model by comparing the model's agreement (or disagreement) with a Bayes-optimal classifier. However, in our work we focus on randomness and noise in the data and how it can effect a model's trust and performance.

Looking at work related to data quality,  Wang et al. \cite{wang1995framework} developd a framework to investigate the research and development of data quality research and concluded that more research is needed to understand data quality and precision.

The impact of noise on machine learning accuracy was explored by \cite{zhu2004class}. They explored noise as being class noise or attribute noise. Attribute noise is noise due to missing or erroneous values. Class noise is noise in classes. They looked at the effect on accuracy from noise by conducting different experiments. However, they only explored classification where in our paper we look at other tasks like regression and explore different models in each of these tasks and finally attempt to create a metric that can quantify model performance.

The concept of uncertainty in machine learning was studied extensively in the survey by \cite{hullermeier2021aleatoric} and \cite{gawlikowski2023survey}. They classify uncertainty as aleatoric and epistemic. Aleatoric uncertainty is due to underlying randomness in the data and epistemic uncertainty is due to lack of observed data. Using the ideas in the survey paper, \cite{staahl2020evaluation} compared four deep learning models: softmax neural network, bayesian neural networks, autoencoders and an ensemble of neural networks in handling uncertainty using entropy \cite{shannon1948mathematical}. However, these works are focused more on the model uncertainty rather than quantifying data uncertainty.

\section{Preliminaries} \label{prelim}

Data is typically classified as being structured or unstructured \cite{majumdar2013big}. Structured data is data that is organized and ordered, like tabular data. Unstructured data includes data with no organization, for example text data. Since unstructured data is converted into structured form, we focus on structured data in this paper. We assume that each column represents a variable and each row represents the record of the data.

We assume that any observed data $Y(t)$ can then be expressed as the sum of the deterministic and non-deterministic components:
\begin{equation} \label{eq:1}
Y(t) = D(t) + E(t)
\end{equation}
In the equation, the observed data is $Y(t)$, $t$ represents the time index or any other relevant index from a finite index set $T$. $D(t)$ is the deterministic component and $E(t)$ is the non-deterministic component.

The objective of this paper is to understand the relationship between the deterministic and non-deterministic components of data and how they influence accuracy of a model. We do not aim to quantify model accuracy as most previous works have done that \cite{hullermeier2021aleatoric, gawlikowski2023survey, staahl2020evaluation}. We hypothesise that non-deterministic components of data can effect the performance of a models in different ways. Any model can easily determine the deterministic component in data, but it is the non-deterministic component that truly effects a model. By quantifying the non-deterministic component of data and using it as a data quality metric, we can better determine the performance of a model.

A non-deterministic function returns different results every time it is called, even when the same input values are provided.

The deterministic component represents the systematic or predictable part of the data. This component can be a function of known predictors, covariates, or any other deterministic relationships and is produced by a well-defined function that always produces the same output when same input values are provided. 

On the other hand, the non-deterministic component of the data is the uncertainty, randomness or unpredictable fluctuations in the data that can be produced by a function that produces different outputs when the input value is same. 

Non-deterministic component of the data can be generated due to many reasons. Random noise or the random nature of the environment can cause randomness and the non-deterministic component in the data. Random noise is often a considered a large component of the noise in data. Random noise is an unavoidable problem. It affects the data collection and data preparation processes, where errors commonly occur. Noise has two main sources: errors introduced by measurement tools and random errors introduced by processing or by experts when the data is gathered.

Throughout this paper we refer to the non-deterministic component as noise even though in some cases it may not be noise as the non-deterministic component may be an inherent property of the data. 

For this purpose, we define a new metric that can quantify the quality of data in terms of how much deterministic and non-deterministic components. We call this new metric \textit{deterministic-to-data ratio ($DDR$)} for a data. 

The inspiration for $DDR$ comes from the concept of signal-to-noise ratio (SNR or S/N). SNR is a measure used in science and engineering that compares the level of a desired signal to the level of background noise. Signal-to-noise ratio is defined as the ratio of the power of a signal to the power of background noise. Depending on whether the signal is a constant or a random variable, the signal-to-noise ratio for random noise becomes the ratio of the square or mean square of the signal to the mean square of the noise. 

Assume $E(t)$ is normally distributed with mean zero and constant variance (homoscedasticity), and that $D(t)$ and $E(t)$ are independent. The $DDR$ of $Y(t)$ can be defined as the ratio of the powers of the deterministic component $D(t)$ and the dataset $Y(t)$:
\resizebox{\linewidth}{!}{
    \begin{minipage}{\linewidth}
\begin{align} \label{eq:2}
    DDR(Y(t)) =\frac{P(D(t))}{P(Y(t))} 
\end{align}
\end{minipage}
}

where $P(D(t)$ and $P(Y(t))$ are the power of $D(t)$ and $Y(t)$ respectively and the power $P: T^{\mathbb{R}} \to \mathbb{R}$ of and any value $X(t)$ as:
\resizebox{\linewidth}{!}{
    \begin{minipage}{\linewidth}
\begin{align} \label{eq:3}
P(X(t)) = 1/|T|\sum_{t = 1}^{|T|}X(t)^2
\end{align}
\end{minipage}
}
Using the definition of power, the $DDR$ can be expressed as (the derivations are in Appendix \ref{appA}):
\resizebox{\linewidth}{!}{
    \begin{minipage}{\linewidth}
\begin{equation} \label{eq:6}
    DDR(Y(t)) \approx \frac{\sum_{t = 1}^{|T|}D(t)^2}{\sum_{t = 1}^{|T|}D(t)^2 + E(t)^2} = \frac{P(D(t))}{P(D(t)) + P(E(t))} 
\end{equation}
\end{minipage}
}
If $DDR \approx 0$, then there is more non-deterministic component in the data and if $DDR \approx 1$ then there is more deterministic component in the data. Hence, when $DDR \approx 0$, then the accuracy of a model would be low as there is more randomness in the data and when $DDR \approx 1$, then the accuracy of a model would be high as there is more deterministic component in the data.

\section{Methodology} 

To understand the performance and trustworthiness of machine learning algorithms, most previous works (e.g. \cite{zhu2004class}) have used synthetic data. The synthetic data are generated by adding noise to a real-world or a generated data set. We call this a top-down approach.

However, we propose a bottom-up approach to generate the data set to measure the performance of a model. In other words, we use a predetermined value for the noise level and modify the real-world or generated data so that the noise level is incorporated into the data set. This gives more control over the noise in the data and standardizes the data set and helps measure the effect of deterministic and nondeterministic components of a data set more fully.

To plot accuracy vs. DDR for several models for a given task, we need to be able to compare accuracy for datasets with different DDRs and models. Typically, one can accomplish this comparison by standardizing the datasets so that for each dataset the mean is $0$ and the variance is $1$. In addition, we would like the accuracy vs. DDR plots to be representative so that the DDRs of the points are uniformly distributed. However, standardization may change the DDRs of datasets so that the distribution of DDRs is no longer uniform.

The noise is added to the features of the data set. The features are the explanatory variables and adding noise to the explanatory or dependent variables affects the independent variable. Since, the objective of any model is to predict the independent variable adding non-deterministic component to features would also effect a model's ability to predict the independent variable.

In the real world splitting the data into deterministic and non-deterministic components is very complicated and the solution may be unknown. There may be infinite possible combinations of deterministic and non-deterministic components. Only an approximate solution may be obtained; e.g. using autoencoders could be used to split deterministic and non-deterministic data. However, for this paper we assume that a data can be split into the deterministic and non-deterministic components. 

To answer the research questions, our aim is to look at how the accuracy changes with changing $DDR$. Using the accuracy-$DDR$ plot, we can obtain useful information that can tell us about the performance of a machine learning model under different levels of non-deterministic components of the data. For the rest of this paper we consider non-determinisitic component as noise.
 
 First, we have to obtain an estimate for the $DDR$ when the dataset is of finite length with zero centered, homoscedastic noise. Equation \ref{eq:1} can be re-written as:
\resizebox{\linewidth}{!}{
    \begin{minipage}{\linewidth}
\begin{equation} \label{eq:7}
Y_i = D_i + E_i
\end{equation}
\end{minipage}
}

where the observed data is $\mathbf{Y_i} = <y_1, ..., y_N>$ with $N$ values. For each value, there is a deterministic component $D_i = <d_1, ..., d_N>$ and non-deterministic component as a random variable or a stochastic process and denote it by $E_i = <e_1, ..., e_N>$.

Equation \ref{eq:6} for $DDR$ can be expressed as:
\resizebox{\linewidth}{!}{
    \begin{minipage}{\linewidth}
\begin{equation} \label{eq:ddr}
DDR(\mathbf{Y}(\mathbf{t})) = \frac{\sum_{i=1}^{N} P(D_i)}{\sum_{i=1}^{N} P(Y_i)}
\end{equation}
\end{minipage}
}

It should be noted that when $i$ is large, then we can get Equation \ref{eq:6}.

There are multiple ways in which $DDR$ can be formulated for a vector of signals. In addition, the $DDR$ is calculated so it is between $0$ and $1$.

The first option to redefine $DDR$ is to define $DDR(Y)$ in terms of each signal $Y_i$. In this case, $DDR(Y)$ would be the mean of sum of $DDR(Y_i)$, but this method would not work when the differences between $DDR(Y_i)$ are too large and could lead to incorrect accuracy-$DDR$ plot. We run into similar issues of taking mean if we consider $\infty$-norm of the $DDR$.  On the other hand, if we consider $DDR$ to be the 1 or 2 norm of the $DDR$s of $DDR(Y_i)$, then $DDR$ of the set may be greater than 1. 

The major problem is how to scale the $DDR$. To scale the $DDR$ when considering the deterministic and non-deterministic component of the data, we propose \textit{DDR-invariant standardization}, a process in which the deterministic and non-deterministic parts for a signal are scaled such that the signal is standardized but the DDR of the signal is preserved. Let $Y(t)$, $T$, $D(t)$, and $E(t)$ be the same as in the definition for signal DDR; $Y^{stand}(t)$ be the DDR-invariant standardized dataset; and $D^{stand}(t)$ and $E^{stand}(t)$, be the objects derived from $Y^{stand}(t)$ in the same way $D(t)$ and $E(t)$ were derived from $Y(t)$. The algorithm for DDR-invariant standardization is presented in Algorithm \ref{alg:ddr_inv}.

\textbf{Theorem}: DDR-Invariant Standardization: $Y^{stand}(t)$ is DDR-invariant standardized if it fulfills the following criteria: 
   \begin{itemize}
       \item $E_{x \in \mathbb{X}}[Y^{stand}(x)] = 0$
       \item $\sigma_{Y^{stand}(t)}^2 = 1$
       \item $DDR(D) = DDR (D^{stand}) = r$
   \end{itemize}    

\RestyleAlgo{ruled}
\begin{algorithm} 
    \caption{DDR-Invariant Standardization}  \label{alg:ddr_inv}
    \KwData{Y(t) = D(t) + E(t)}
    \KwResult{Standardized $D(t) \& E(t)$}
    \For {Y(t)} {Calculate $\alpha, \beta, and \sigma_{E^{stand}(t)}^2 $ using equation \ref{eq: 15} - \ref{eq: 17} \\
    Determine $D^{stand}(t)$ using equation \ref{eq: 11}}
\end{algorithm}

Using approximations, $\alpha, \beta \& \sigma$ can be approximated as (see Appendix \ref{appB}:
\resizebox{\linewidth}{!}{
    \begin{minipage}{\linewidth}
\begin{align} 
& \alpha \approx \frac{\sqrt{r}}{S_{D(t)}} \label{eq: 15} \\
& \beta = - \frac{\overline{D(t)}}{S_{D(t)}}\times \sqrt{r}  \label{eq: 16} \\
& \sigma_{E^{stand}(t)}^2 \approx 1 - r  \label{eq: 17} 
\end{align}
\end{minipage}
}

where $S_{D(t)}$ is the sample standard deviation of $D(t)$ and $\overline{D(t)}$ is the sample mean of $D(t)$.

To obtain $DDR(Y)$, we define $DDR$ for sets of signals "at a lower level" by expanding the definition of power for a signal to a set of signals. We obtain a representative accuracy for a given data set $DDR = R$, we average the accuracy for a set of data sets generated by hit-and-run sampling \cite{francesc_font-clos_2021}. 

Hit-and-run sampling belongs to a class of procedures called symmetric mixing algorithms that recursively generates a sequence of points all within a given region. These points have the property that when the initial point is uniformly distributed within the region, all of them are. Moreover, the last generated point is asymptotically independent of the first point as the sequence grows in length. The general mixing algorithm is given in Algorithm \ref{alg:mixing} in Appendix \ref{appC}. Hit-and-run algorithm is the same algorithm with $D = \{d \in \mathbb{R}^n| \norm{d} = 1\}$ \cite{smith1984efficient}. 

In hit-and-run sample analysis $n$ -tuples of $DDR$ $<r_1, ..., r_n>$, each with the same $DDR = R$, from the convex polytope defined by the relation between $R$, $r_1, ..., r_n$ and by the constraints on $r_1, ..., r_n$ are generated. Ideally, we would sample $n$-tuples of $DDR$ so that the resulting $n$-tuples are approximately uniformly distributed. However, under the 2-norm definition, and after a process we term \textit{DDR-invariant standardization}, the relation between $R$, $r_1, ..., r_n$ becomes $n\times R^2 = r_1^2 + ... + r_n^2$. In this case, hit-and-run sampling guarantees that the $n$-tuples of squares of $DDR$s $<r_1^2, ..., r_n^2>$, and not the corresponding $n$-tuples of $DDR$s, are approximately uniformly distributed. 

\begin{figure}[htbp]
\centering
\includegraphics[width=7.5cm]{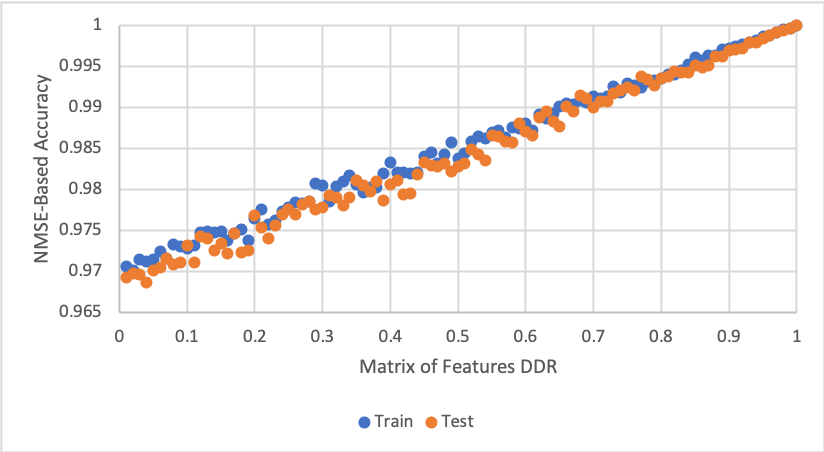}
\caption{NMSE-Based Accuracy of Ordinary Least Squares Regression vs. Matrix of Features DDR} \label{Ordinary_Least_Squares_Regression}
\end{figure} 

\begin{figure}[htbp]
\centering
\includegraphics[width=7.5cm]{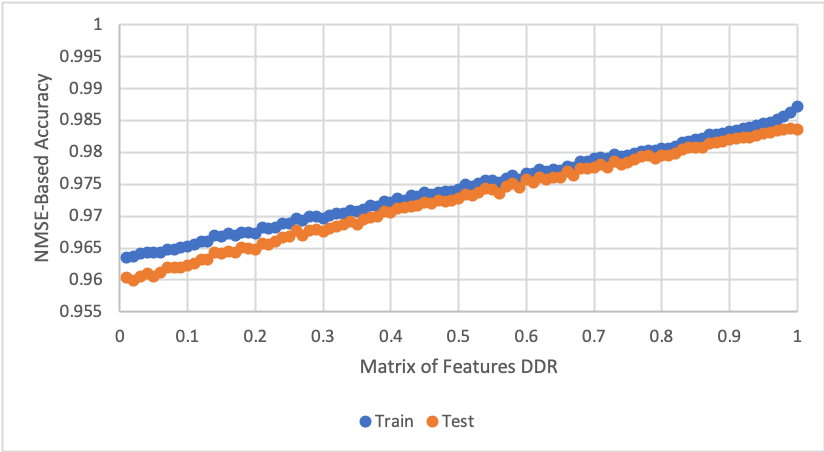}
\caption{NMSE-Based Accuracy of Decision Tree Regression vs. Matrix of Features DDR} \label{Decision_Tree_Regressor}
\end{figure} 

\begin{figure}[htbp]
\centering
\includegraphics[width=7.5cm]{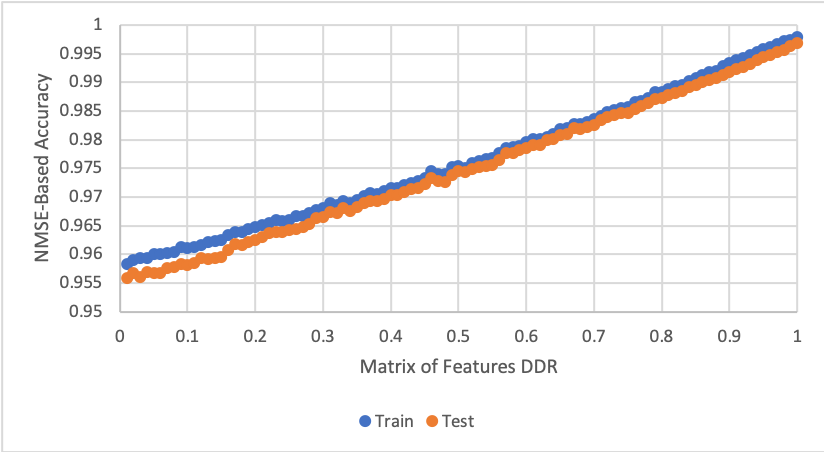}
\caption{NMSE-Based Accuracy of KNearest Neighbors Regression vs. Matrix of Features DDR} \label{KNearest_Neighbors_Regressor}
\end{figure}

\begin{figure}[htbp]
\centering
\includegraphics[width=7.5cm]{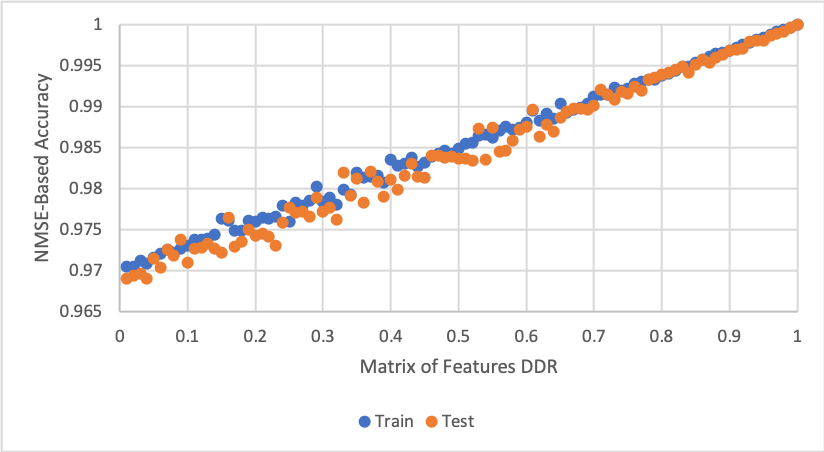}
\caption{NMSE-Based Accuracy of Linear Support Vector Regression vs. Matrix of Features DDR} \label{Linear_Support_Vector_Regressor}
\end{figure}

\begin{figure}[hbp]
\centering
\includegraphics[width=7.5cm]{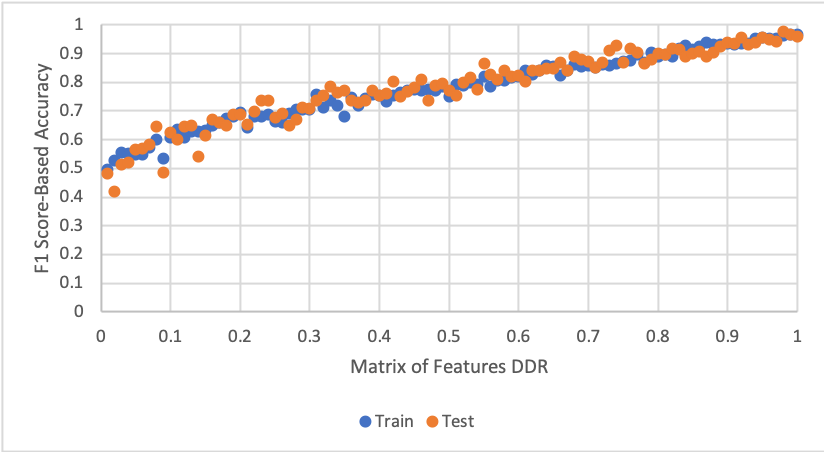}
\caption{F1-Based Accuracy of Binary Logistic Regression 2-Class Classification vs. Matrix of Features DDR} \label{Binary_Logistic_Regression_Classifier}
\end{figure}

\begin{figure}[htbp]
\centering
\includegraphics[width=7.5cm]{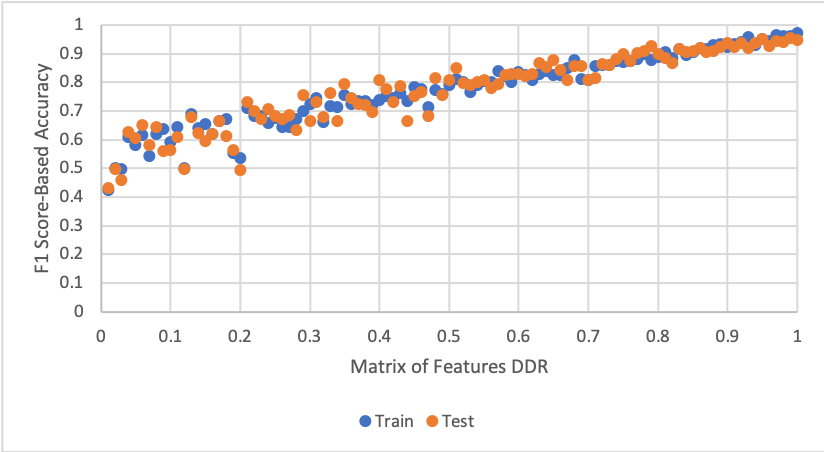}
\caption{F1-Based Accuracy of Decsision Tree Classification vs. Matrix of Features DDR} \label{Decision_Tree_Classifier}
\end{figure}

\begin{figure}[htbp]
\centering
\includegraphics[width=7.5cm]{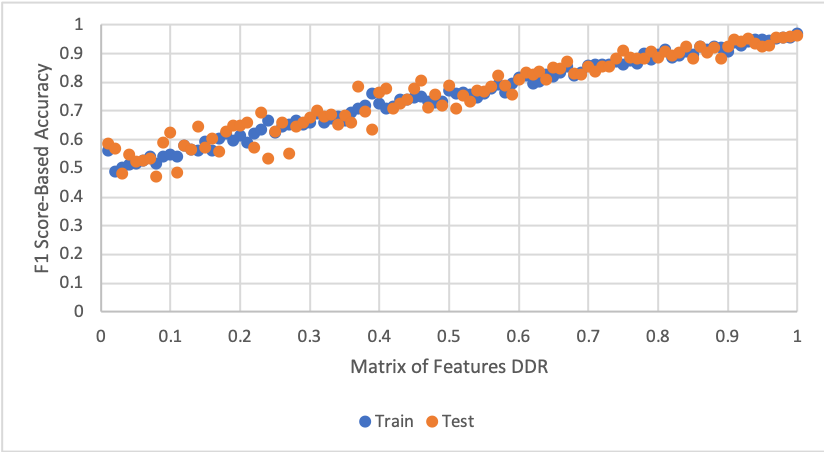}
\caption{F1-Based Accuracy of KNearest Neighbors Classification vs. Matrix of Features DDR} \label{KNearest_Neighbors_Classifier}
\end{figure}

\begin{figure}[htbp]
\centering
\includegraphics[width=7.5cm]{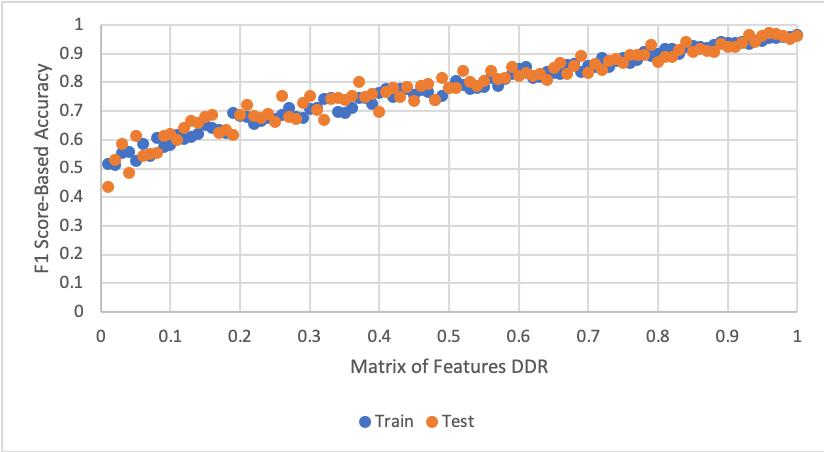}
\caption{F1-Based Accuracy of Linear Support Vector Classsification vs. Matrix of Features DDR} \label{Linear_Support_Vector_Classifier}
\end{figure}

\begin{figure}[htbp]
\centering
\includegraphics[width=7.5cm]{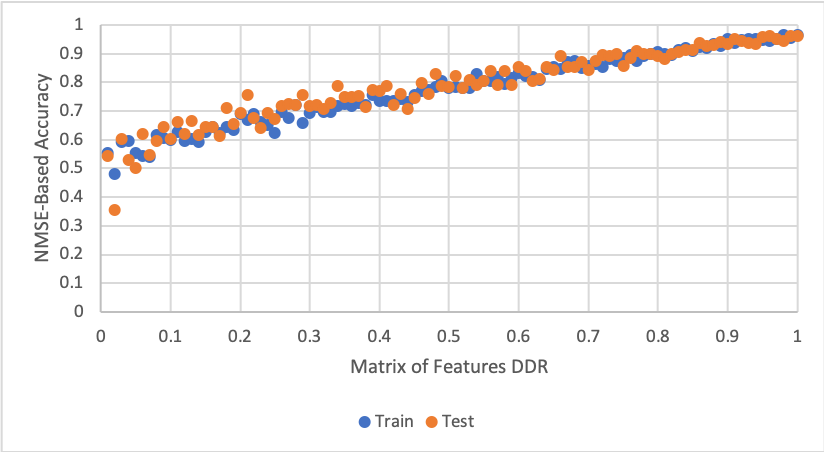}
\caption{F1-Based Accuracy of Multi-Layer Perceptron Classification vs. Matrix of Features DDR} \label{Multi_Layer_Perceptron_Classifier}
\end{figure}

To generate the accuracy-$DDR$ plot, we consider many different types of models. To measure accuracy, we look at different metrics. The metrics are dependent on the type of task and can be found as labeled in the plot. We look mainly at regression and classification.

\section{Model Performance Metric}

In this paper a new metric, the trustworthiness portfolio $p$, was defined. The trustworthiness portfolio measures the performance of a model. As a single-number measure of the reliability of a model M on a given noisy data set Y, for a given model and the deterministic component, trustworthiness portfolio can be obtained from the accuracy-$DDR$. 

The trustworthiness portfolio should be a metric that looks at the change in performance of a model when the non-deterministic component (or the noise) changes. Ideally, performance should not change when $DDR$ changes. Since $DDR$ and accuracy are normalized between $0$ and $1$, the maximum value of the trust should be 1. Hence, the trustworthiness portfolio is between $0 \leq DDR \leq 1\ \ \forall DDR$. When data set has $DDR = 0$ then the data has a high non-deterministic component level and there is no reliable way of making predictions without a lot of pre-processing. 

Hence, a simple way of defining trustworthiness portfolio is:
\begin{align} \label{eq:trust}
    p_{M,Y} = accuracy \times DDR
\end{align}
The above definition of metric is for a single point. For a specific model and dataset, we can use the accuracy-DDR plot to measure the true performance of a model under uncertain conditions. The true performance of a model is defined as follows:

\begin{align} \label{eq:perf}
    p_M = \int_1^0 accuracy(DDR) d(DDR)
\end{align}

where $accuracy(DDR)$ is the function from the accuracy-DDR plot. It should be noted that $p_M$ is equivalent to the area-under-the-curve of the accuracy-DDR plots. When $p_M$ is close to 1, then model has high performance and can perform well under uncertain conditions. This happens when the model accuracy does not change with changing DDR which is ideally how a model should perform. When $p_M<1$ when model's accuracy changes with changing DDR and the lower the value, worst is the performance of the model under uncertain conditions.

\section{Experimental Design \& Result} \label{exp}

In this section, we look at the experiments that we performed to create the accuracy-DDR plots. The experiments were done using Python and implemented Pytorch \cite{NEURIPS2019_9015} and Scikit-learn \cite{scikit-learn}. Pytorch was generally used for data generation and implementing different machine learning models. Scikit-learn was used to get basic machine learning models, along with other places used to get widely accessible versions of common and uncommon models. Our code can be found at \href{https://github.com/ucinAI800/pyModelingMLPerformance}{Github}

We collected scores for ten types of machine learning models on multiple appropriate
datasets with varying DDRs of feature matrix. The models were not altered in any way from where we obtained them unless specified in the code, this includes no hyperparameter tuning being performed. We included ten supervised learning model types, which consisted of five regression model types and five binary classification model types. The regression model types included a linear regression model type, Ordinary Least Squares Regressor (OLSRR); Decision Tree Regressor (DTR); a support vector regression model type, Linear Support Vector Regressor (LSVR); a nearest neighbors regression model type, K-Nearest Neighbors Regressor (KNNR). The binary classification model types included a linear binary classification model type, Binary Logistic Regression Classifier (BLRC); Decision Tree Classifier (DTC); a support vector binary classification model type, Linear Support Vector Classifier (LSVC); a nearest neighbors binary classification type, K-Nearest Neighbors Classifier (KNNC); and a neural network binary classification model type, Multilayer Perceptron Classifier (MLPC).

We measured the results of each model's accuracy based on it's type. All five regression models' accuracy was measured using normalized mean square error (NMSE)-based accuracy. The five binary classification models' accuracy were measured using F1-Score.

The plots of the results can be found in Figures \ref{Ordinary_Least_Squares_Regression}, \ref{Decision_Tree_Regressor}, \ref{KNearest_Neighbors_Regressor}, \ref{Linear_Support_Vector_Regressor}, \ref{Binary_Logistic_Regression_Classifier} for the regression algorithms and \ref{Binary_Logistic_Regression_Classifier}, \ref{Decision_Tree_Classifier}, \ref{KNearest_Neighbors_Classifier}, \ref{Linear_Support_Vector_Classifier}, \ref{Multi_Layer_Perceptron_Classifier} for the different classification algorithm. The plots show the accuracy for both the training and testing samples.

The results of the plots show a clear trend. As DDR increases, accuracy also decreases which translates to as there is more deterministic component in the data, higher is the accuracy. For the accuracy-DDR plot for the regression algorithm, the accuracy-DDR relationship is almost linear. Except for multi-layer perception (MLP), accuracy does not change significantly with DDR. This indicates that non-deterministic component in the data has very little effecft on MLP. Hence, it is the most reliable algorithm compared to all the other regression algorithms. 

When it comes to classification, the trend is also linear except for the low DDR values, where the relationship is slightly logarithmic. In addition, the linear relationship gradient is also more gradual compared to regression plots. Here again MLP has the best performance compared to the other classification algorithm. 

\begin{figure}[htbp] 
\centering
\includegraphics[width=7.5cm]{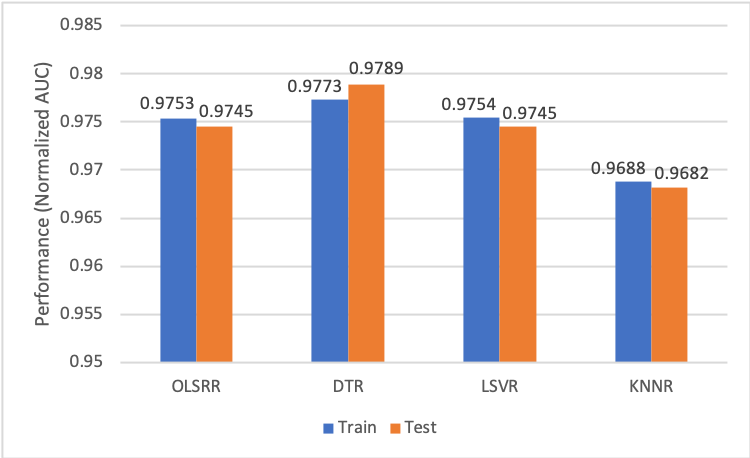}
\caption{Performance Results for Regression Models} \label{Regression_Performance_Bar_Plot}
\end{figure}

\begin{figure}[htbp] 
\centering
\includegraphics[width=7.5cm]{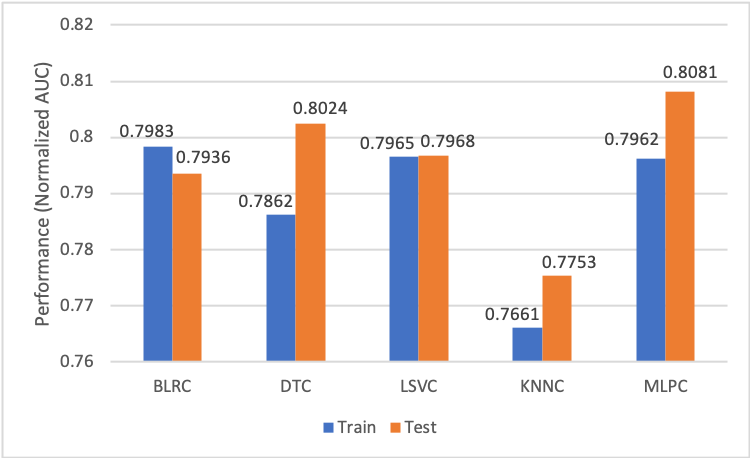}
\caption{Performance Results for Classification Models} \label{Classification_Performance_Bar_Plot}
\end{figure}

Finally, in Figures \ref{Regression_Performance_Bar_Plot} and \ref{Classification_Performance_Bar_Plot} we show the trustworthiness portfolio of a model using equation \ref{eq:perf} referred to as normalized AUC on the y-axis. For both regression and classification models KNNC models perform the worst with uncertain conditions. DTR has the best performance in regression. MLPC classification models perform the best under uncertainity which verify prior research and their popularity in usage as classification models.

\section{Conclusion} \label{conc}

In summary, we proposed a metric called $DDR$, which we showed to be a valid measure of the non-deterministic component in a dataset and the trustworthiness portfolio for a model that incorporates $DDR$ to measure the performance. Our results also in line with the claims from other research like Zhu et al. \cite{zhu2004class} who showed that the accuracy of a decision tree classifier increases as the noise level of a data set decreases. But we have gone beyond the work in previous research by creating a framework that can consider different machine learning types, model types, and evaluation metrics. 

We also proposed a new bottom-up approach of generating synthetic data and \textit{DDR-invariant-standardization} to standardize the data for measuring the true performance of a model. 

Our method here is preliminary and there are multiple avenues for future research. Our method could be applied to the new concept of data-centric AI which involved data engineering and developing data-driven AI models \cite{zha2023data0, zha2023data, khang2023data, whang2023data}. The work described here could also be a step towards explainable AI and developing better understanding of AI models \cite{bertossi2020data}, especially under uncertain conditions. Other potentials for future research include further generalizing the results of this study by including clustering model types, ensemble models, or complex neural networks, using multiple appropriate evaluation metrics per learning type, using multiple appropriate data sets per model type with varying numbers of features, using class noise or both feature and class noises, or modeling uncertainty and noise using multiple kinds of feature noises such as heteroscedastic noise and/or uniform noise. 

We also would like to extend the definition for trustworthy portfolio, e.g. by proposing a Bayesian-based definition that also incorporates other properties of the dataset in addition to uncertainty.

\bibliography{sample-base}
\newpage

\onecolumn

\title{Supplementary Material}
\maketitle

\appendix
\section{Appendix A} \label{appA}

In this section, we derive the equations for $DDR$.

\begin{align} \label{eq:3a}
P(X(t)) = 1/|T|\sum_{t = 1}^{|T|}X(t)^2
\end{align}

Using the definition of power, the $DDR$ can be expressed as:

\begin{align} \label{eq:4}
& DDR(Y(t)) = \frac{\frac{\sum_{t = 1}^{|T|}D(t)^2}{|T|}}{\frac{\sum_{t = 1}^{|T|}Y(t)^2}{|T|}} = \nonumber \\
& \frac{\sum_{t = 1}^{|T|}D(t)^2}{\sum_{t = 1}^{|T|}D(t)^2 + \sum_{t = 1}^{|T|}D(t)\times E(t) + \sum_{t = 1}^{|T|}E(t)^2}
\end{align}

If $|T|$ is very large, then:

\begin{align} \label{eq:5}
& \sum_{t = 1}^{|T|}D(t)\times E(t) \approx |T|\times \mathbb{E}[D(t)E(t)] \nonumber \\ 
& = |T| \times \mathbb{E}[D(t)]\mathbb{E}[E(t)] = 0
\end{align}

Since $D(t)$ and $E(t)$ are independent. then:

\begin{equation} \label{eq:6a}
    DDR(Y(t)) \approx \frac{\sum_{t = 1}^{|T|}D(t)^2}{\sum_{t = 1}^{|T|}D(t)^2 + E(t)^2} = \frac{P(D(t))}{P(D(t)) + P(E(t))} 
\end{equation}

\section{Appendix B} \label{appB}

For large $|T|$, the expectation can be approximated by the sample mean, and the following:

\begin{align}  \label{eq: 9}
    \frac{\sum_{t = 1}^{|T|}Y^{stand}(t)}{|T|} \approx 0
\end{align}

\begin{align}  \label{eq: 10}
    \frac{\sum_{t=1}^{|T|} Y^{stand}(t)^2}{|T|} \approx 1
\end{align}

Next, we assume that $D^{stand}(t)$ is a linear transformation of $D(t)$:

\begin{align} \label{eq: 11}
    D^{stand}(t)=\alpha \times D(t) + \beta 
\end{align}
where $\alpha$ is positive and $\beta \in \mathbb{R}$. Upon solving for $\alpha$, $\beta$, and $\sigma_{E^{stand}(t)}^2$, we found that:

\begin{align}
\alpha = \frac{\sqrt{r}}{\sigma_{D(t)}} \\
\beta = - \frac{\mathbb{E}[D(t)]}{\sigma_{D(t)}}\times \sqrt{r} \\
\sigma_{E^{stand}(t)}^2 = 1 - r
\end{align}

\section{Appendix C} \label{appC}

\RestyleAlgo{ruled}
\begin{algorithm} 
    \caption{Mixing Algorithm}  \label{alg:mixing}
    \KwData{Input a bounded $k$-dimensional surface $S \subset \mathbb{R}^n$, where $k \leq n$, and number of iterations $N$}
    \KwResult{Output a set of $N + 2$ points $\{X_0, X_1, ..., X_{N+1}\}$ uniformly distributed in $S$}
    \textbf{Initialization:} Randomly uniformly pick a starting point $X_0 \in S$ \\
    \For {i in [N]} {Generate a random direction $d$ uniformly distributed over a \textit{direction set} $D \subseteqq \mathbb{R}^n$.
    
    Find the \textit{line set} $L = S \cap \{x|x=x_0+\lambda d, \lambda \textrm{ a real scalar}\}$.
    
    Generate a random point $X_{i+1}$ uniformly distributed on $L$.}
\end{algorithm}

\section{Appendix D} \label{appD}

\begin{table*}[htbp]
\caption{Experiment Results - Regression Performance (Normalized AUC)}
\begin{center}
\begin{tabular}{|c|c|c|c|c|c|}
\hline
\textbf{Method} & \textbf{Train} & \textbf{Test}\\
\hline
Ordinary Least Squares Regression & $0.975342$ & $0.974469$\\
\hline
Decision Tree Friedman 1 Regression & $0.977254$ & $0.978903$\\
\hline
Linear Support Vector Random Regression & $0.975373$ & $0.974527$\\
\hline
K Nearest Neighbors Friedman 1 Regression & $0.968825$ & $0.968244$\\
\hline

\end{tabular}
\label{Regression_Performance_Table}
\end{center}
\end{table*}

\begin{table*}[htbp]
\caption{Experiment Results - Classification Performance (Normalized AUC)}
\begin{center}
\begin{tabular}{|c|c|c|c|c|c|}
\hline
\textbf{Method} & \textbf{Train} & \textbf{Test}\\
\hline
Binary Logistic Regression Random 2-Class Classification & $0.798254$ & $0.793556$\\
\hline
Decision Tree Random 2-Class Classification & $0.786196$ & $0.802385$\\
\hline
Linear Support Vector Random 2-Class Classification & $0.796526$ & $0.796809$\\
\hline
K Nearest Neighbors Random 2-Class Classification & $0.766083$ & $0.77528$\\
\hline
Multi-Layer Perceptron Random 2-Class Classification & $0.796177$ & $0.80809$\\
\hline

\end{tabular}
\label{Classification_Performance_Table}
\end{center}
\end{table*}

\end{document}